\documentclass[letterpaper, 10 pt, conference]{ieeeconf}
\usepackage{}  

\usepackage{amssymb}
\usepackage{amsmath}
\usepackage{algorithmic}
\usepackage{algorithm}
\usepackage{graphicx,epstopdf}
\usepackage{url}
\usepackage{gensymb}
\usepackage{cite,color}
\usepackage[tight,footnotesize]{subfigure}

\newcommand{\be}{\begin{eqnarray}}
\newcommand{\ben}{\begin{eqnarray*}}
\newcommand{\en}{\end{eqnarray}}
\newcommand{\enn}{\end{eqnarray*}}

\IEEEoverridecommandlockouts                              

\title{\LARGE \bf An  Efficient Generation Method based on Dynamic Curvature of the Reference Curve for Robust Trajectory Planning }

\author{ \authorblockN{Yuchen Sun,~ Dongchun Ren,~ Shiqi Lian,~ Mingyu Fan,~ and Xiangyi Teng}
\authorblockA{Section of Self-driving Vehicle\\
Meituan Group,
Beijing, China 100102\\
Email: \{sunyuchen03, lianshiqi, rendongchun, fanmingyu, tengxiangyi\}@meituan.com}
\and
}

\begin{document}

\maketitle
\thispagestyle{empty}
\pagestyle{empty}

\begin{abstract}
Trajectory planning is a fundamental task on various autonomous driving platforms, such as social robotics and self-driving cars. Many trajectory planning algorithms use a reference curve based Frenet frame with time to reduce the planning dimension. However, there is a common implicit assumption in classic trajectory planning approaches, which is that the generated trajectory should follow the reference curve continuously. This assumption is not always true in real applications and it might cause some undesired issues in planning. One issue is that the projection of the planned trajectory onto the reference curve maybe discontinuous. Then, some segments on the reference curve are not the image of any part of the planned path. Another issue is that the planned path might self-intersect when following a simple reference curve continuously. The generated trajectories are unnatural and suboptimal ones when these issues happen. In this paper, we firstly demonstrate these issues and then introduce an efficient trajectory generation method which uses a new transformation from the Cartesian frame to Frenet frames. Experimental results on a simulated street scenario demonstrated the effectiveness of the proposed method.
\end{abstract}

\section{INTRODUCTION}

Trajectory planning is used to create a safe and smooth trajectory for an agent (e.g., a robot or a self-driving car) from one state (including the location, the velocity, the acceleration, the orientation and so on) to another state under certain constraints or cost functions \cite{TrajPlanning2017}. There are many ways to plan a trajectory for an intelligent agent and it is a common practice to use a reference curve, which is generated off-line, as the target to find the output trajectory. In this way, the trajectory planning algorithm only need to follow the given reference curve and does not require the information of map like signs, turns and so on.  To achieve optimality and time efficiency, many trajectory planning algorithms \cite{ref1,ref2,ref3}  are developed in Frenet frames with time to reduce the planning dimension with the help of a reference curve. In a Frenet frame, finding the optimal trajectory is essentially a three dimensional constrained optimization problem.


There are roughly two types of trajectory planning algorithms: the direct methods \cite{Spatiotemporaliros2009,ref11} that use trajectory sampling or lattice search to find the optimal trajectory and the path-speed decoupled methods \cite{decouple2015,baiduEMplanner2018} that optimize path and speed separately. In both of the methods, all obstacles and environment information are projected onto the lane center based Frenet frames for each candidate lane, and the drivable area is bounded by lane boundaries. When using the Frenet frame, a projection operator which identifies the corresponding projection point on the reference curve is always required and used as the transformation from the Cartesian frame to Frenet frames. The most widely used projection operator is defined as the nearest point on the reference curve, which is also the orthogonal projection point. Given a reference curve, there are two sub-tasks in trajectory planning that a projection operator is necessary, which are:
\begin{itemize}
\item {\bf Reference curve following:} One has to determine whether a candidate trajectory is actually following the lane by checking its projection on the reference curve. \\
\item {\bf Trajectories generation:}  One need to generate as many as possible candidate trajectories that follows the reference curve for subsequent planning and optimization.
\end{itemize}
However, under certain circumstances, the projection of a candidate trajectory onto the reference curve maybe discontinuous \cite{ref2,ref3} , which is commonly ignored in real applications. Some continuous segments on the reference curve could not be mapped back onto the generated trajectory and make the first sub-task difficult.  Moreover, trajectories generated by previous methods might self-intersect when continuously following a simple reference curve. The generated trajectories might be unnatural and suboptimal trajectories for the second sub-task. These discussed issues motivate us to develop novel a trajectory generation method that could produce reasonable candidate trajectories under most of the circumstances. In this paper, based on our theoretical result on trajectory following,  the derived method could generate trajectories that guarantee smooth, anomaly free, and follow the reference curve always.

\begin{figure}
	\centering
    \includegraphics[width=8.5cm]{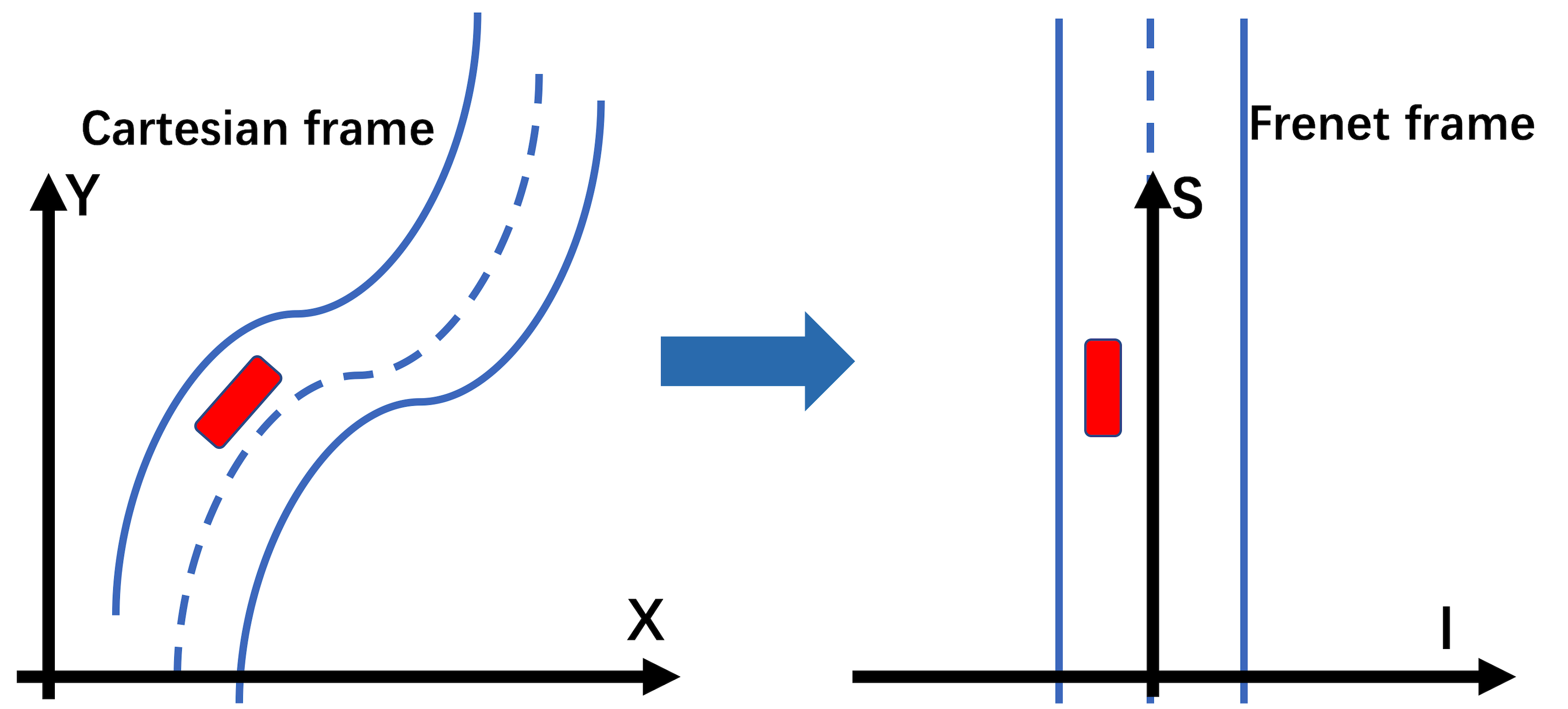}
	\caption{An illustration of the transformation from Cartesian frame to Frenet frames}\label{counterExamples}
\end{figure}

\section{RELATED WORK}

The past decades have witnessed a great progress in trajectory planning. Many researches have been proposed and studied \cite{ref2}, \cite{ref7,ref8,ref9} that they can generate global trajectories connecting a start and a possibly distant end state. Some approaches for trajectory planning follow a discrete sampling or a road map-based scheme \cite{ref5} \cite{ref6}, which firstly sample multiple rows of points and then generate a finite set of trajectories with cost values, typically by differential functions that describe vehicle dynamics and kinematics. A candidate trajectory is selected if and only if it corresponds to the minimal cost value.

Kelly and Nagy \cite{ref10} proposed an efficient path planning algorithm which employs polynomials to ensure the smooth variation of the curvature. Based on this method, a new approach proposed in \cite{ref11} firstly samples points along the reference lane and then connects the sampled points using polynomials. In this way, all generated candidate trajectories conform to the lane shape. However, these methods use Cartesian coordinate system rather than the Frenet frames. Therefore, the planning dimension is relatively high and the optimization is complicated.

To address the issue, Werling et al. \cite{ref2} used the lane center line based Frenet frames  to decouple the lateral and longitudinal motion. This strategy makes the reference curve coincide with the lane shape. Based on Frenet frame, the  generation of the lateral and longitudinal motions can be achieved separately by using quintic polynomials versus time, which ensures continuous acceleration. Given the advantages of the Frenet frames, many trajectory planning methods \cite{ref3,ref12} are proposed based on it and achieve a roaring success in addressing various planning problems in real applications. However, most of the methods ignore an intrinsic problem in  the transformation from Cartesian frame to Frenet frames, which  may cause unnatural and suboptimal trajectories for the downstream tasks.  In this paper, based on our theoretical result on trajectory following,  the proposed method could generate candidate trajectories following the reference curve and guarantee smoothness and anomaly free.

 \begin{figure}
	\centering
    \includegraphics[width=7cm]{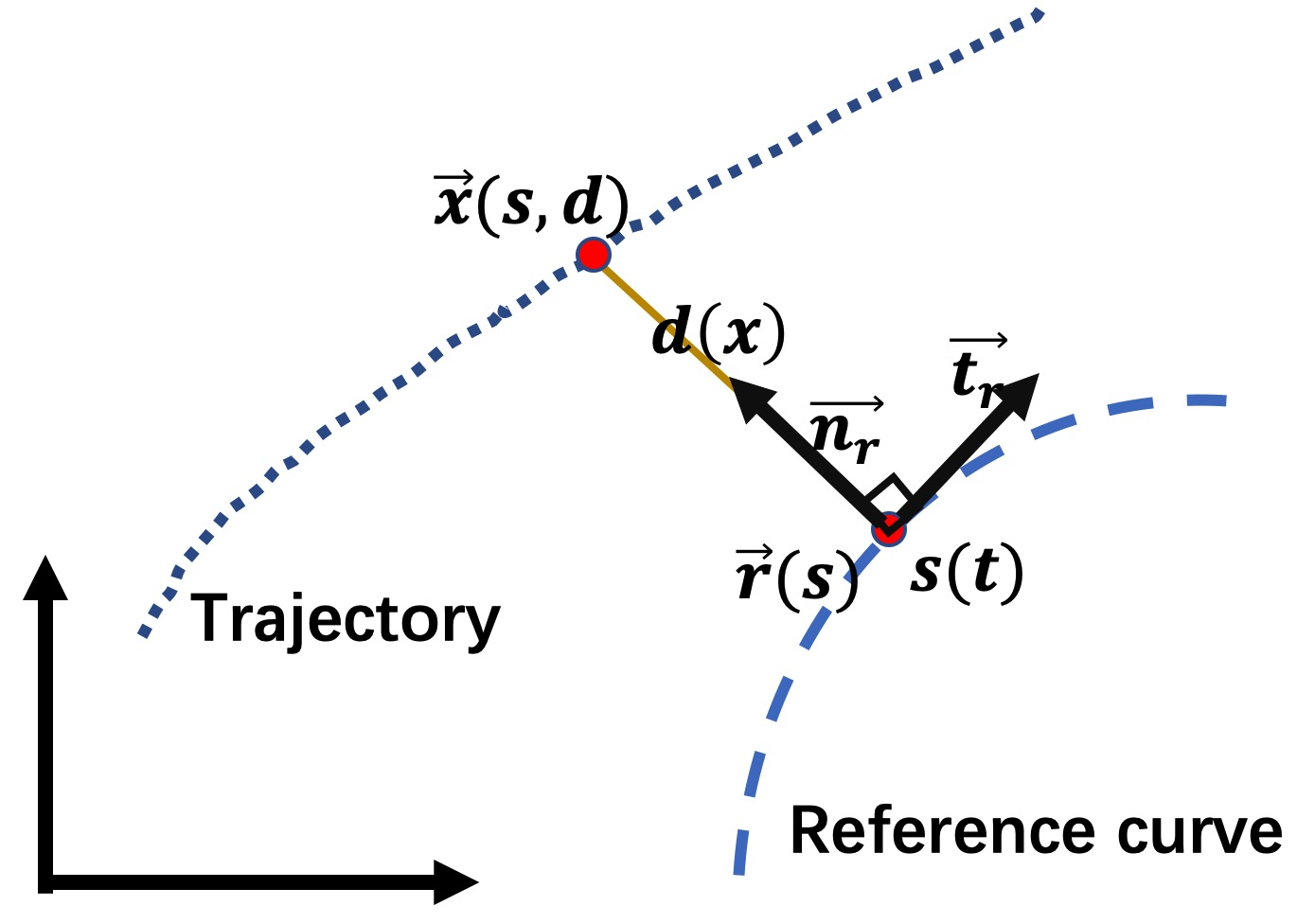}
	\caption{The representation of a candidate trajectory in a Frenet frame}\label{counterExamples}
\end{figure}

\section{Preliminaries and Our Theoretical Results}

\begin{figure*}[htbp]
\centering
 \subfigure[discontinuous]{\includegraphics[width=7cm]{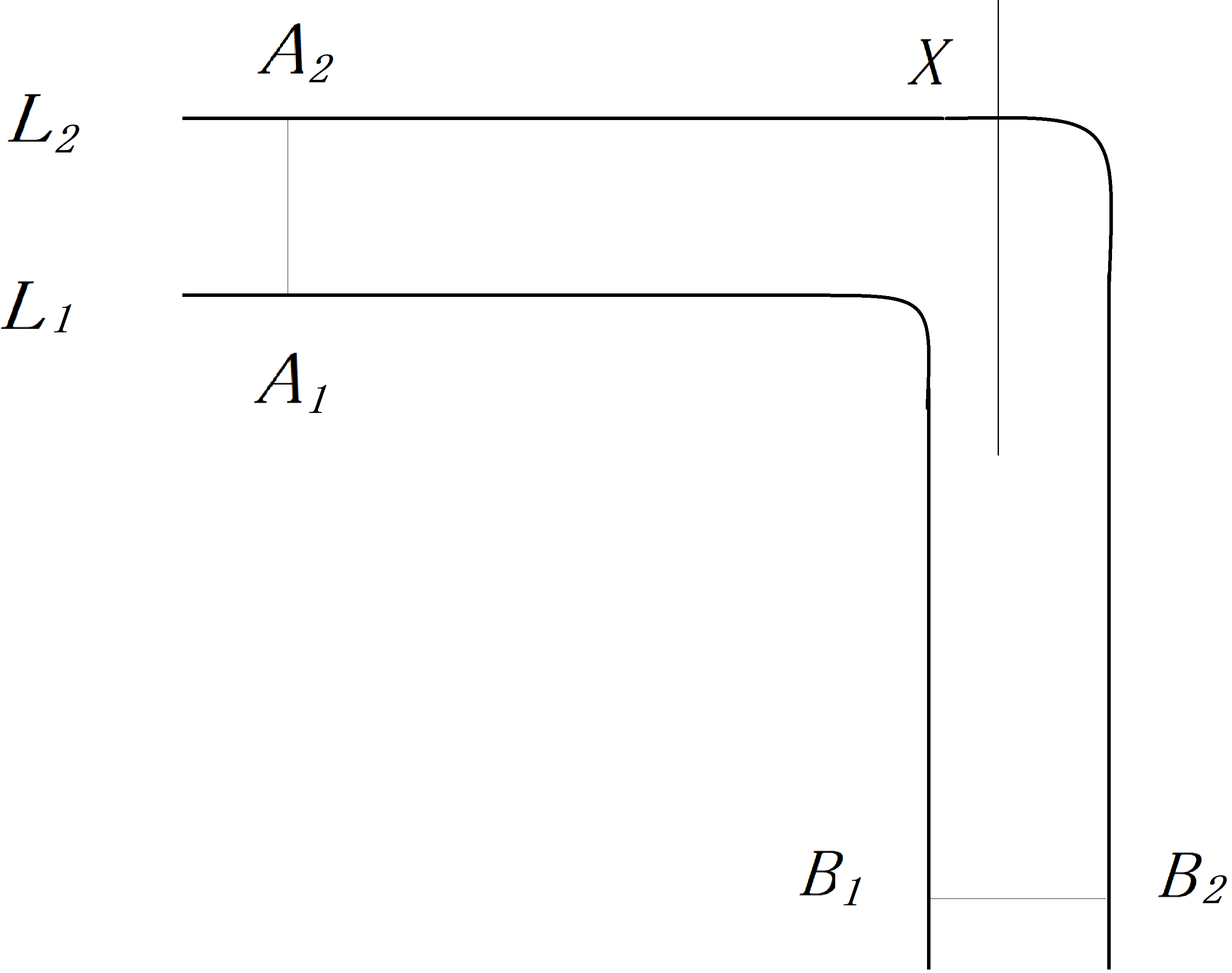}}
 \subfigure[self-interact]{\includegraphics[width=7.2cm]{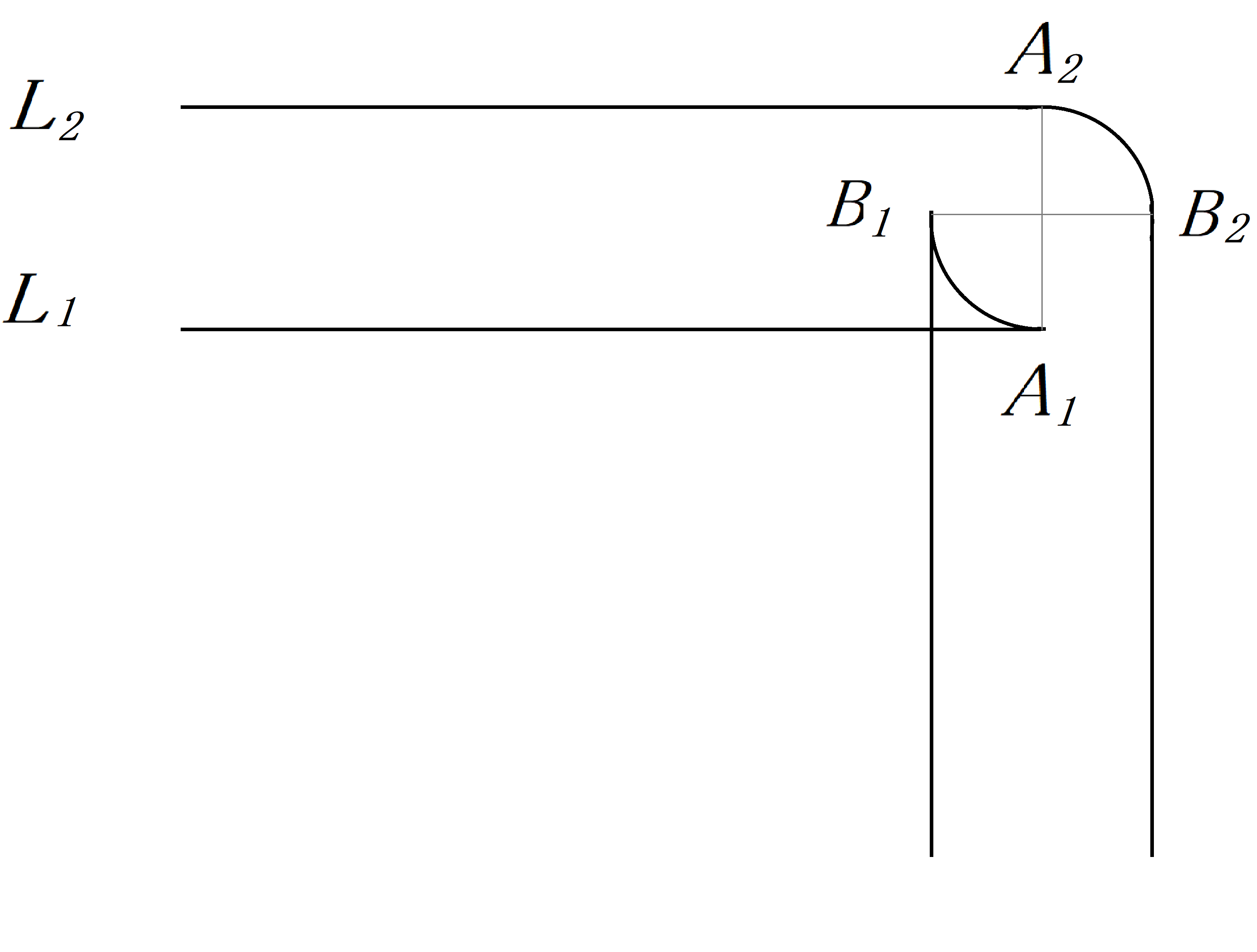}}
\caption{The issues of discontinuous projection operator $\textbf{p}$, (a) multiple outputs for a single $\vec{x}$, and the projection of $L_1$ onto $L2$ is discontinuous, (b) a suboptimal trajectory when following the reference curve faithfully. $L_1$ is a candidate trajectory and $L_2$ is the reference curve}\label{counterExamples}
 \end{figure*}

\subsection{The Frenet Frame}

A reference curve is always defined by parametric curves \cite{ref13}\cite{ref14} such as the spline curve. When $s(t)$ (or $s$ for short) is the modular length of time $t$, we assume the vector function $\vec{r}(s)$ denotes a moving point on the reference curve $L$ in Euclidean space. Let $\vec{t}_r(s)$ and $\vec{n}_r(s)$ are the unimodular tangent vector and normal vector at $\vec{r}(s)$, respectively. Then, $\left\{\vec{t}_r(s),  \vec{n}_r(s)\right\}$ formulates a dynamic orthogonal coordinate system. For any Cartesian coordinates $\vec{x}$, assuming $\vec{r}(s)$ is the root point of $\vec{x}$,  we have
\be
\vec{x}(s,k_x,d_x) =\vec{r}(s) + k_x \vec{t}_r (s) + d_x \vec{n}_r (s),
\en
where $k_x = k_x(t)$ and $d_x = d_x(t)$ are the horizontal and perpendicular offsets, respectively. And the frenet coordinates of $\vec{x}$ can be denoted by $(s,k_x,d_x)$. Figure 1 is an example of a Frenet frame. To simplify the expression, one common practice is to use the nearest point to $\vec{x}$ on $L$ as the root point, such that $k_x=0$. Then the Frenet coordinate of $\vec{x}$ can be denoted by $(s(t),d_x(t))$ and the corresponding projection operator $\textbf{p}$ of $\vec{x}$ onto $L$ is $s(t)$.
That is to say, the projector ${\bf p}$ satisfies:
\be \label{projector}
{\bf p}(\vec{x}) = s(t).
\en


\subsection{The Projection Operator}

Considering $L_2$ be the reference curve and the $L_1$ is a candidate trajectory of a moving agent, there are two implicitly ideal properties/desirable assumptions of the projection operator $\textbf{p}$ for trajectory planning tasks \cite{ref2}, which are summarized as below.
\begin{itemize}
\item $\textbf{p}$ is a continuous and smooth projector such that the image ({\bf math}) of a continuous segment on $L_1$ should be a continuous interval on $L_2$.
\item The moving agent on the candidate trajectory $L_1$ should be actually following the reference curve $L_2$, which can be mathematically represented as $\langle \dot{{\bf p}}(\vec{x}),\vec{t}_r(s)\rangle >0$.
\end{itemize}

However, the definition of the projection operator $\textbf{p}$ given in (\ref{projector}) is  incomplete and the neither of the two desired properties is guaranteed in most of the previous motion planning studies. That means that it may have several outputs at some point $\vec{x}$ when following the given reference curve $L_2$. Fig. \ref{counterExamples}(a) presents a reference curve with a 90 degrees turn. When a virtual agent on $L_1$ moves into the turn with a reasonable large distance to the reference curve $L_2$, the trajectory $L_1$ should be contained in the available set for optimization. If the projection operator $\textbf{p}$ is continuous, the image of $\textbf{p}$ should contain all segment between $A_2$ and $B_2$. As can be seen, the point $X$ on $L_2$ is not the projection of any point from $L_1$ and this violates the assumption that $\textbf{p}$ is continuous. Fig. \ref{counterExamples}(b) presents an example in path following. The reference curve has a rotation arc whose radius is $R$. The virtual agent is following $L_2$ continuously with the constraint that the lateral distance is constant $d_x(t) = 2R$. When the agent moves from $B_1$ to $A_1$, the direction of its movement on the candidate trajectory is opposite with the direction of projection point on the reference curve.

\newtheorem{theorem}{Theorem}[section]
\newtheorem{proposition}{Proposition}[section]
\newtheorem{corollary}{Corollary}[section]
\newtheorem{definition}{Definition}[section]
\newtheorem{lemma}{Lemma}[section]

\subsection{Our Theoretical Results}

To address the intrinsic issues of the definition of the projection operator $\textbf{p}$,  the curvature $\kappa$ is required to develop a novel transformation, to replace the original projection operator $\mathbf{p}$, from Cartesian coordinates to Frenet coordinates. Our method is based on the following theoretical result.

\begin{theorem} \label{th1}
Let $L_2$ be the reference curve and $\vec{x}$ be any point (an agent) on the candidate trajectory $L_1$. $\vec{r}(t)$ is the nearest point on $L_2$ to $\vec{x}$ and $\kappa d_x(t)<1$ , then $\vec{r} - \vec{x}$ is perpendicular to the tangent $\vec{t}_r(t)$ and the agent is actually following $L_2$ at the same direction.
\end{theorem}

The theorem \ref{th1} indicates that the projection point $\vec{r}(t)$ follows reference curve $L_2$  at the same direction if and only if $\kappa d_x(t)<1$.  Moreover, $\vec{r}(t)$ will jump and be discontinuous  on $L_2$ at these points where the curvature $\kappa$ is large.  If a method can handle the discontinuous points properly and guarantee $\kappa d_x(t)<1$, it can generate the smoothness of the generated trajectories which follow the reference curve at the same direction.

For the convenience the downstream optimizers, it is a common practice to discretize the reference curve in the form of vectors.  The discretization is realized through point sampling on the reference curve. The sampling points on the curve is connected into a polygonal line, where the sampling points are sequenced in the order of their distance to one end point of the curve. The obtained polygonal line is used to represent the original curve. To avoid the intractable cases such as sharp angles and fractals, the following empirical conditions should be satisfied in the sampling process:
\begin{itemize}
\item The distance between two adjacent samples should be extremely smaller than $1/\kappa$.
\item The angle between two adjacent sample points should not exceed 10$\degree$.
\item The distances between every pair of adjacent samples are identical.
\end{itemize}

It should be noted  that the polygonal line has finite length due to the limitations of sampling size. As a result, the obtained polygonal line that is built by line segments is not enough to ensure that the normal line at each point on the reference curve go through of it. To deal with this, the linear extension strategy \cite{ref16} is adopted to extend the polygonal line. Specifically,  the first and last segments on the polygonal line are extended to rays with infinite length.

\begin{algorithm}
\renewcommand{\algorithmicrequire}{\textbf{Input:}}
\renewcommand{\algorithmicensure}{\textbf{Output:}}
  \caption{The Projection Operator: TransF-$\kappa$}
  \label{alg1}
  \begin{algorithmic}[1]
  \REQUIRE
  The polygonal line in the form of vectors: $U= \lbrace L_c: (x_c, y_c)|c=1,2,...,M \rbrace$ , which is formulated by sampling points from the reference curve $L$ ($M$ denotes the sampling size);
the Cartesian coordinates of a point  $A: (x_A, y_A)$.
  \ENSURE The Frenet coordinate of $A: (s_A, d_A)$
  \STATE Compute the distances between $A$ and all points in $U$.
  \STATE Identify all the points in $U$ that have the identical and the nearest distance to $A$, which form a subset $N$.
   \STATE Choose the point $L_m$ such that $m=\arg \max c : L_c \in N$, i.e., the biggest index from subset $N$.
  \IF{$L_m$ is not an end point}
  \STATE $v_+ = \langle \overrightarrow{L_m A},\overrightarrow{L_m L_{m+1}} \rangle$ {\color{red}\# where $\overrightarrow{X_1 X_2} = X_2-X_1$ is the difference of the coordinates of two points}
  \STATE $v_- = \langle \overrightarrow{L_m A},\overrightarrow{L_m L_{m-1}} \rangle$  {\color{red}\# where $\langle \cdot, \cdot \rangle$ is the inner product of two vectors}
  \IF{$v_- < v_+$}
  \STATE $m=m+1$
  \ENDIF
  \STATE ($s_A, d_A$)={\bf AffineTrans}$(A,m,U)$ {\color{red} \# Algo 2}
  \ELSE
  \STATE ($s_A, d_A$)={\bf ParallelTrans}$(A,m,U)$   {\color{red} \#  Algo 3}
  \ENDIF
  \RETURN $A: (s_A, d_A)$
  \end{algorithmic}
\end{algorithm}


\begin{algorithm}
\renewcommand{\algorithmicrequire}{\textbf{Input:}}
\renewcommand{\algorithmicensure}{\textbf{Output:}}
  \caption{Affine Transform}
  \label{alg2}
  \begin{algorithmic}[1]
  \REQUIRE
  The polygonal line in the form of vectors: $U= \lbrace L_c: (x_c, y_c)|c=1,2,...,M \rbrace$  ($M$ denotes the sampling size); the index $m$ of the reference piece;  The Cartesian coordinates of a point  $A: (x_A, y_A$)
  \ENSURE The Frenet coordinate of $A: (s_A, d_A)$
  \STATE Get the heading vector $\vec{b_1}$ of the angular bisector at the junction $L_{m-1}$.
  \STATE Get the heading vector $\vec{b_2}$ of the angular bisector at the junction $L_{m}$.
  \IF{$\vec{b_1}$ is parallel to $\vec{b_2}$}
  \STATE ($s_A, d_A$)=$ParallelTrans(A,m,U)$
  \ELSE
  \STATE  $s_{m-1} = \sum_{i=1}^{m-1} \Arrowvert \overrightarrow{L_{i-1} L_i}\Arrowvert$
  \STATE Calculate the intersection point $O$ of the two angular bisectors, which go through points $L_{m-1}$ and $L_{m}$ with headings $\vec{b_1}$ and $\vec{b_2}$, respectively.
  \STATE Calculate the intersection point $P$ of line $\overrightarrow{OA}$ and $\overrightarrow{L_{m-1}L_m}$.
  \STATE $s_A = s_{m-1} + \Arrowvert \overrightarrow{L_{m-1} P}\Arrowvert$
  \STATE $d_A=$ distance from $A$ to the line $L_{m-1}L_m$.
  \ENDIF
  \RETURN ($s_A, d_A$)
  \end{algorithmic}
\end{algorithm}

\begin{algorithm}
\renewcommand{\algorithmicrequire}{\textbf{Input:}}
\renewcommand{\algorithmicensure}{\textbf{Output:}}
  \caption{ Parallel Transform}
  \label{alg3}
  \begin{algorithmic}[1]
  \REQUIRE
  The polygonal line in the form of vectors: $U= \lbrace L_c: (x_c, y_c)|c=1,2,...,M \rbrace$  ($M$ denotes the sampling size); the index $m$ of the reference piece;  The Cartesian coordinates of  $A: (x_A, y_A$)
  \ENSURE The Frenet coordinate of $A: (s_A, d_A)$
  \IF{$m =$ 0}
  \STATE Denote  $\overrightarrow{L_0L_1}$ as $\vec{E}$.
  \STATE Set $s_A=0$.
  \STATE Denote  $\overrightarrow{L_0 A}$ as $\vec{V}$
  \ELSE
  \STATE Denote  $\overrightarrow{L_{m-1}L_m}$ as $\vec{E}$
  \STATE Set $s_A = \sum_{i=1}^{m-1} \Arrowvert \overrightarrow{L_{i-1}L_i} \Arrowvert$
  \STATE Denote  $\overrightarrow{L_{m-1} A}$ as $\vec{V}$
  \ENDIF
  \STATE Let $\vec E_{\perp}$ be a nonzero orthogonal vector to $\vec E$ such that $ \langle \vec E_{\perp}, \vec E\rangle =0$.
  \STATE $s_A=s_A+ \vec V \cdot \vec E / \Arrowvert \vec E \Arrowvert$.
  \STATE $d_A=  | \langle \vec V, \vec E_{\perp}\rangle / \Arrowvert \vec E_{\perp} \Arrowvert |$
  \RETURN ($s_A, d_A$)
  \end{algorithmic}
\end{algorithm}

\section{The Proposed Method}

In this section, we propose a new efficient trajectory generation method such that the generated candidate trajectories are smooth and guarantees to follow the reference curve at the same direction. The method is developed based on a new projection operator,  TransF-$\kappa$, from the Cartesian frame to Frenet frames.

\subsection{Our Projection Operator: TransF-$\kappa$}

Based on the theorem \ref{th1} and sampling requirements, it is intuitive and natural to project a moving point onto the nearest point on the reference curve $L$. However, there may be several nearest projection points with the identical distance. In our projection operator: TransF-$\kappa$, we choose the projection point with the biggest $s$ to ensure that the moving point is actually following at the same direction with $L$, where $s$ denotes the distance on curve from the projection point to the start point on $L$. {\bf Algorithm \ref{alg1}} shows the details of TransF-$\kappa$, the transformation from a Cartesian coordinates to the  Frenet coordinate of an arbitrary point.

In {\bf Algorithm \ref{alg1}} lines 1-2, we first pick out all points on $U$ which are the nearest to $A$ and form a set containing all such points. Then we choose the last one, i.e. one whose index is the biggest. At line 3, This point with the biggest $s$ is used and the output trajectory is shorten. With the point $L_m$, lines 5-9 decide A should be projected to the former or latter adjacent piece to $L_m$. The condition at line 7 identifies which side $A$ should be projected on. The inverse image of two pieces are separated by the angular bisector at $L_m$.  Finally, lines 10-12 are used to calculate the projection point. The Affine Transform given in {\bf Algorithm \ref{alg2}} is applied on the pieces of $U$ without linear extension while the Parallel Transform given in {\bf Algorithm \ref{alg3}} is employed on the extended pieces of $U$. One significant difference between these two operators is that the inverse image of common pieces is bounded by two angular bisectors while the extended pieces is bounded by only one angular bisector. The calculation approaches of the projection results are based on these two situations. The details of Affine Transform operator and Parallel Transform operator are given in {\bf Algorithm \ref{alg2}} and {\bf \ref{alg3}}, respectively.

Further discussions about  {\bf Algorithm \ref{alg2}} and {\bf \ref{alg3}} should be made for clarification. As can be seen, both Affine and Parallel Transform are approximation to the transform function Eq.  (1) in Section III. The reason why we need the approximation methods are that the polygonal line $U$ as a sampled approximation for reference curve $L$ is no longer smooth and thus there is no curvature, tangent vector and so on. Therefore, two approximation operators are used to make the lateral distance $d$ vary continuously when the point $A$ moves cross an angular bisector. In other words, they guarantee that point $A$ on an angular bisector has the same projection point on $U$ no matter it is projected to the former or latter piece. However, since they are approximation methods, some error may occur. An apparent error is that the created coordinate system may not be orthogonal to the reference curve. That is why we introduce some sampling conditions which should be satisfied in Section III.C. Those conditions ensure that the approximation errors are small enough to be ignored and make them work well in real-world applications.

\begin{algorithm}
\renewcommand{\algorithmicrequire}{\textbf{Input:}}
\renewcommand{\algorithmicensure}{\textbf{Output:}}
  \caption{ A Commonly-used Continuous Candidate Trajectory Generation Method ({\bf May cause self-interaction!})}
  \label{alg4}
  \begin{algorithmic}[1]
  \REQUIRE
  The polygonal line in the form of vectors: $U= \lbrace L_c: (s_c, d_c=0)|c=1,2,...,M \rbrace$  ($M$ denotes the sampling size); Vectors of the lateral boundaries at the sample points: $B_c = \lbrack a_c, b_c \rbrack$, $c=1,2,...,M$.
  \ENSURE A candidate trajectory $T$ consisted of vectors of $M$ points: $T = \lbrace T_c: (s_c', d_c')|c=1,2,...,M \rbrace$
  \STATE $i = 0$
  \STATE $T= \varnothing$
  \WHILE {$i< M$ }
  \STATE $i=i+1$
  \STATE $d_i'$= randomly pick a number between $a_i$ and $b_i$
  \STATE $s_i'=s_i$
  \STATE $T=T \cup (s_i',d_i')$
  \ENDWHILE
  \RETURN A candidate trajectory $T$
  \end{algorithmic}
\end{algorithm}

\begin{algorithm}
\renewcommand{\algorithmicrequire}{\textbf{Input:}}
\renewcommand{\algorithmicensure}{\textbf{Output:}}
  \caption{The Proposed Trajectory Generation Method}
  \label{alg5}
  \begin{algorithmic}[1]
  \REQUIRE
  An $n$-order differentiable reference curve $L$;  The polygonal line in the form of vectors: $U= \lbrace L_c: (s_c, d_c=0)|c=1,2,...,M \rbrace$;  The candidate trajectory of $M$ points: $T = \lbrace T_c: (s_c', d_c')|c=1,2,...,M \rbrace$ which is generated by {\bf Algo \ref{alg4}}.
  \ENSURE A renewed candidate trajectory $T'$ from $T$: $T''$= $\lbrace T_c''(s_c'', d_c'')|c=1,2,...,M \rbrace$
  \STATE $i = 0$
  \STATE $T''=\varnothing$
  \WHILE {$i< M$ }
  \STATE $i=i+1$
  \STATE Calculate $\kappa$ = curvature of $L$ at $L_i$($s_i$). 
  \IF {$\kappa d_i' < 1$}
  \STATE $T''=T'' \cup (s_i',d_i')$
  \ELSE
  \STATE Calculate the Euclidean coordinates of Frenet coordinates $(s_i',d_i')$, which is denoted as $(x_i,y_i)$.  
  \STATE Calculate the Frenet coordinates of $(x_i,y_i)$ referring to $U$,  $(s_i'',d_i'')$ = TransF-$\kappa$ (U,$(x_i,y_i)$). {\color{red} \# Algo1}
  \STATE $T''=T'' \cup (s_i'',d_i'')$
  \STATE $i = min \lbrace i|s_i'>s_i''\rbrace$
  \ENDIF
  \ENDWHILE
  \RETURN A renewed trajectory $T''$ from $T'$
  \end{algorithmic}
\end{algorithm}

\subsection{Our Trajectory Generation Algorithm}

Trajectory generation is a key step in trajectory planning in dynamic and unknown environments. The downstream trajectory optimizer \cite{ref15} may fail to create an acceptable trajectory plan that obeys traffic rules and without misleading directions (such as self-interactions) when the input candidate trajectories are sloppy. {\bf Algorithm \ref{alg4}} is a common framework for trajectory generation. Many sophisticated generation methods can be extended from {\bf Algorithm \ref{alg4}}. Without loss of generality, we choose {\bf Algorithm \ref{alg4}} as the basis of our generation method.

As discussed in Section III, Algorithm \ref{alg4} and the related sophisticated generation methods  will produce candidate trajectories with misleading directions and self-interaction when the curvature is large and the gap of the corridor boundaries is wide.  Typical examples can be found in Fig. \ref{counterExamples}.  

To ensure the candidate trajectory is actually following at the same direction as the reference curve, our method introduce the curvature $\kappa$ that $\kappa d_x (t) < 1$ as an additional restriction condition to repair the generated trajectories by {\bf Algorithm \ref{alg4}}.   {\bf Algorithm \ref{alg5}} is our proposed trajectory generation method which uses the outputs of {\bf Algorithm \ref{alg4}}.  

The inputs of  {\bf Algorithm \ref{alg5}} include the $n$-order differentiable reference curve $L$. The curvature $\kappa$ at any points on $L$ can be calculated by $\kappa = \langle  \dot{t_r}, n_r \rangle$, where $t_r = \vec{t_r}(s)$ and $n_r = \vec{n_r}(s)$ are  the uni-modular tangent and normal vectors at $\vec r_s$ on $L$, respectively (the same notations as in Section III.A).  The input of  {\bf Algorithm \ref{alg5}} $T$ is an output of  {\bf Algorithm \ref{alg4}}, which might consist of misleading directions and self-interactions.  The output of  {\bf Algorithm \ref{alg5}} is a renewed candidate trajectory $T'$, which is repaired from $T$.  At the line 5 of {\bf algorithm 5}, the curvature at the sampling point $L_i$ is calculated. If $\kappa d_i' < 1$, it is safe to include the point $T_i:(s_i',d_i')$ into the candidate trajectory.  Otherwise, at lines 9-12, the point have to be transformed into the Cartesian frame and then transformed back into the Frenet frame.  Because several nearest points to $(x_i,y_i)$  might be identified in $U$, $(s_i'',d_i'')$ is different with $(s_i',d_i')$. In this way, the segment on reference curve $L$ with high curvature $\kappa$ is jumped over and the obtained new point is safe to be included into $T''$. 

\section{SIMULATION STUDY AND DISCUSSION}

\begin{figure*}[htbp] 
\centering
\subfigure[]{\includegraphics[width=4.1cm]{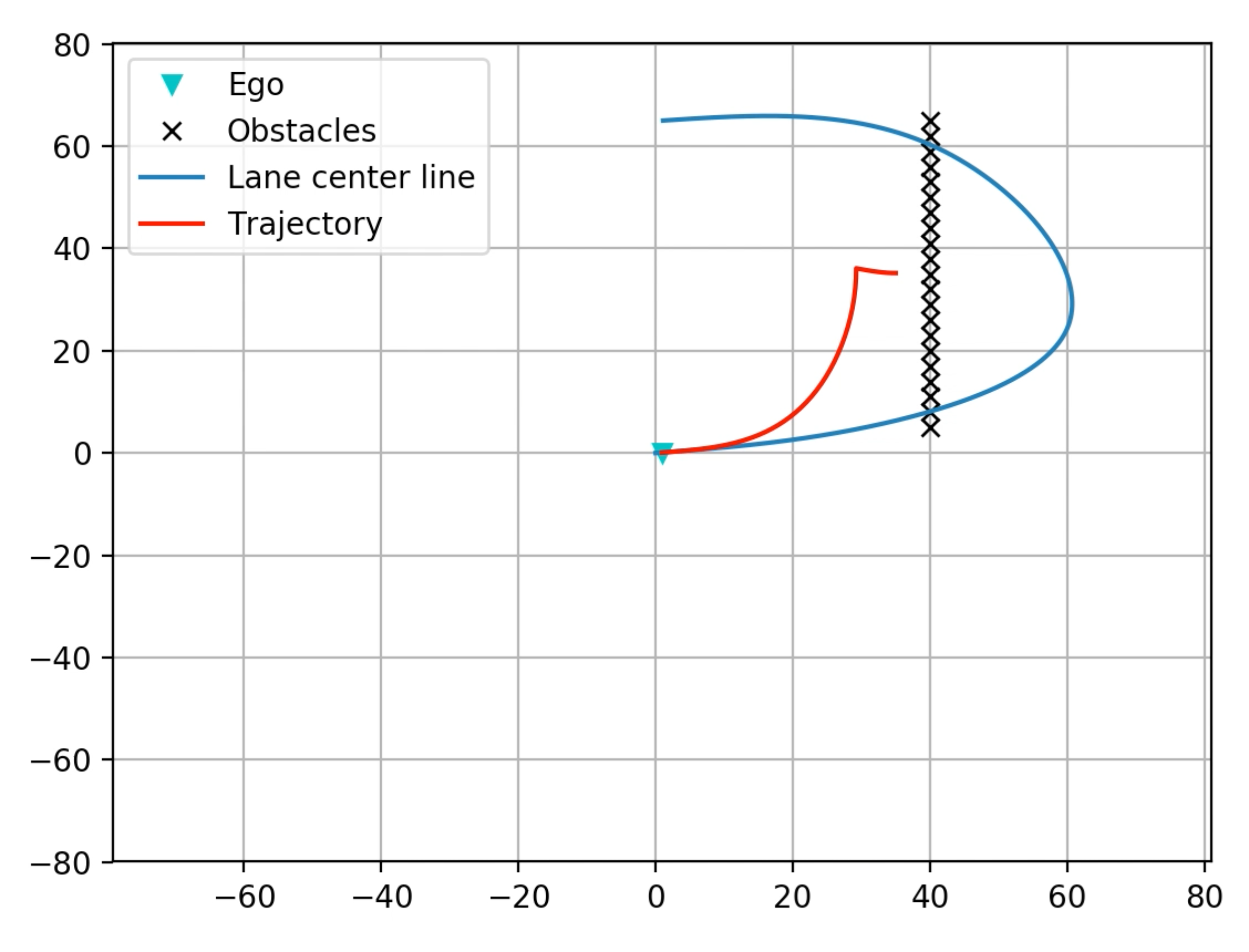} }
\subfigure[]{\includegraphics[width=4.1cm]{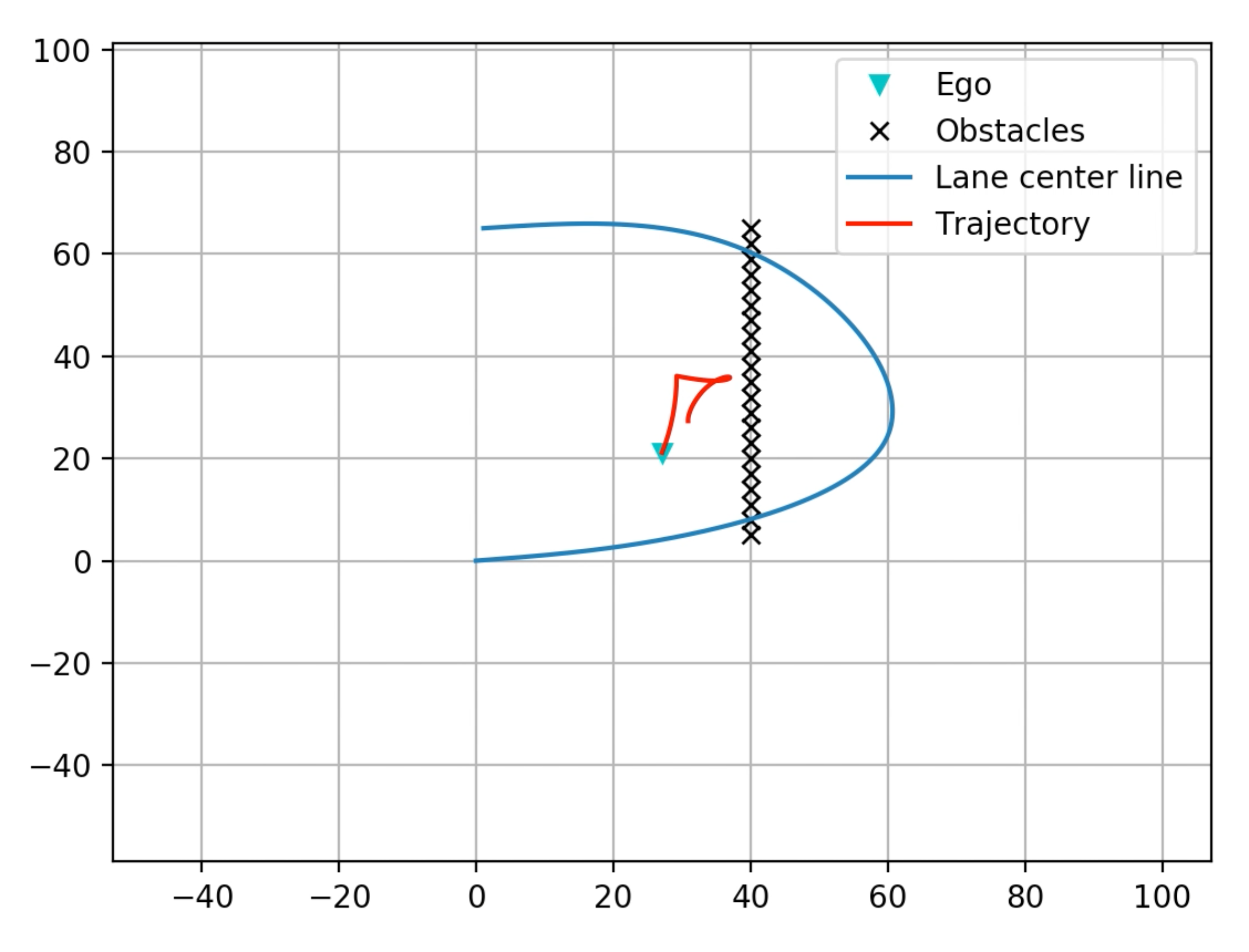} }
\subfigure[]{\includegraphics[width=4.1cm]{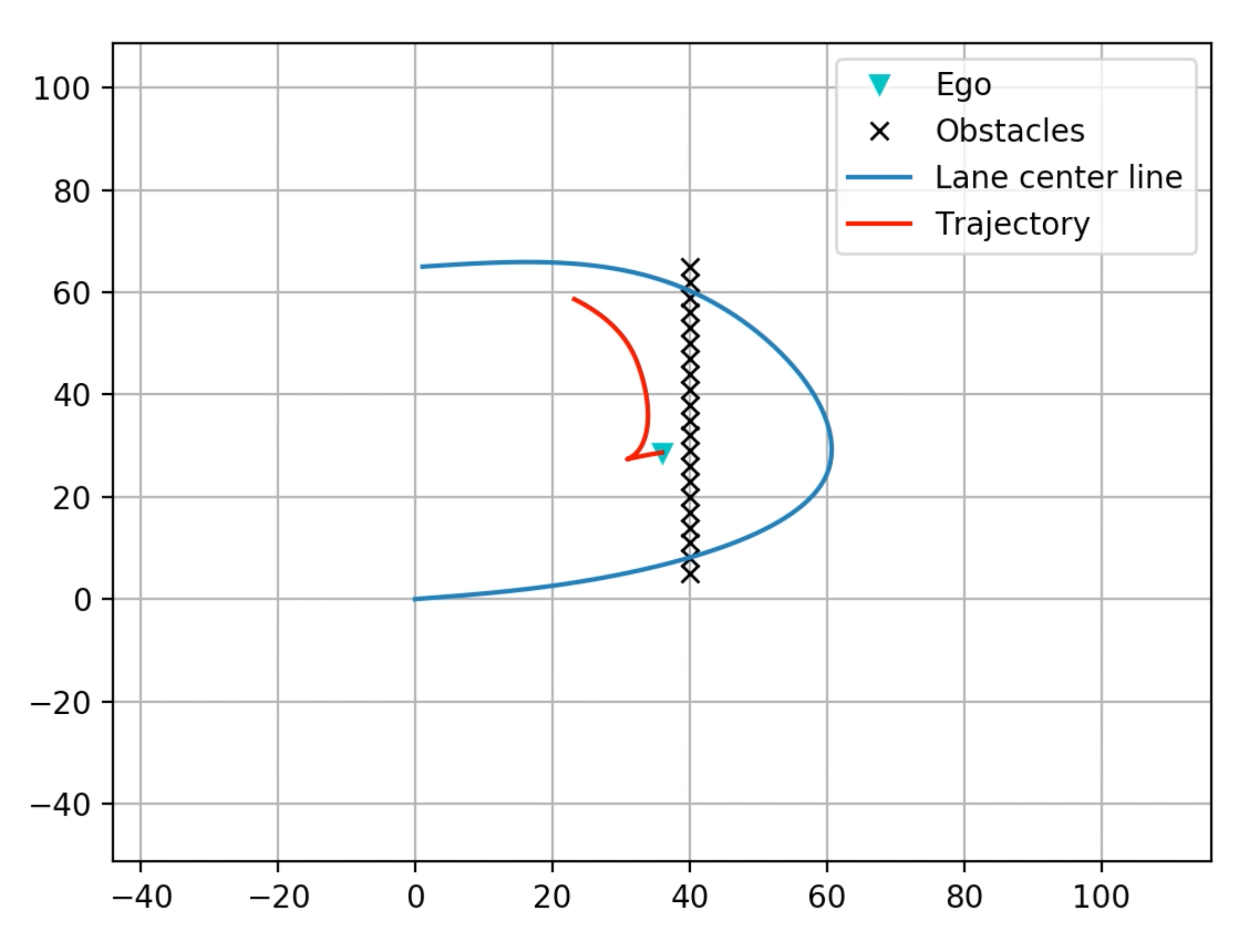} }
\subfigure[]{\includegraphics[width=4.1cm]{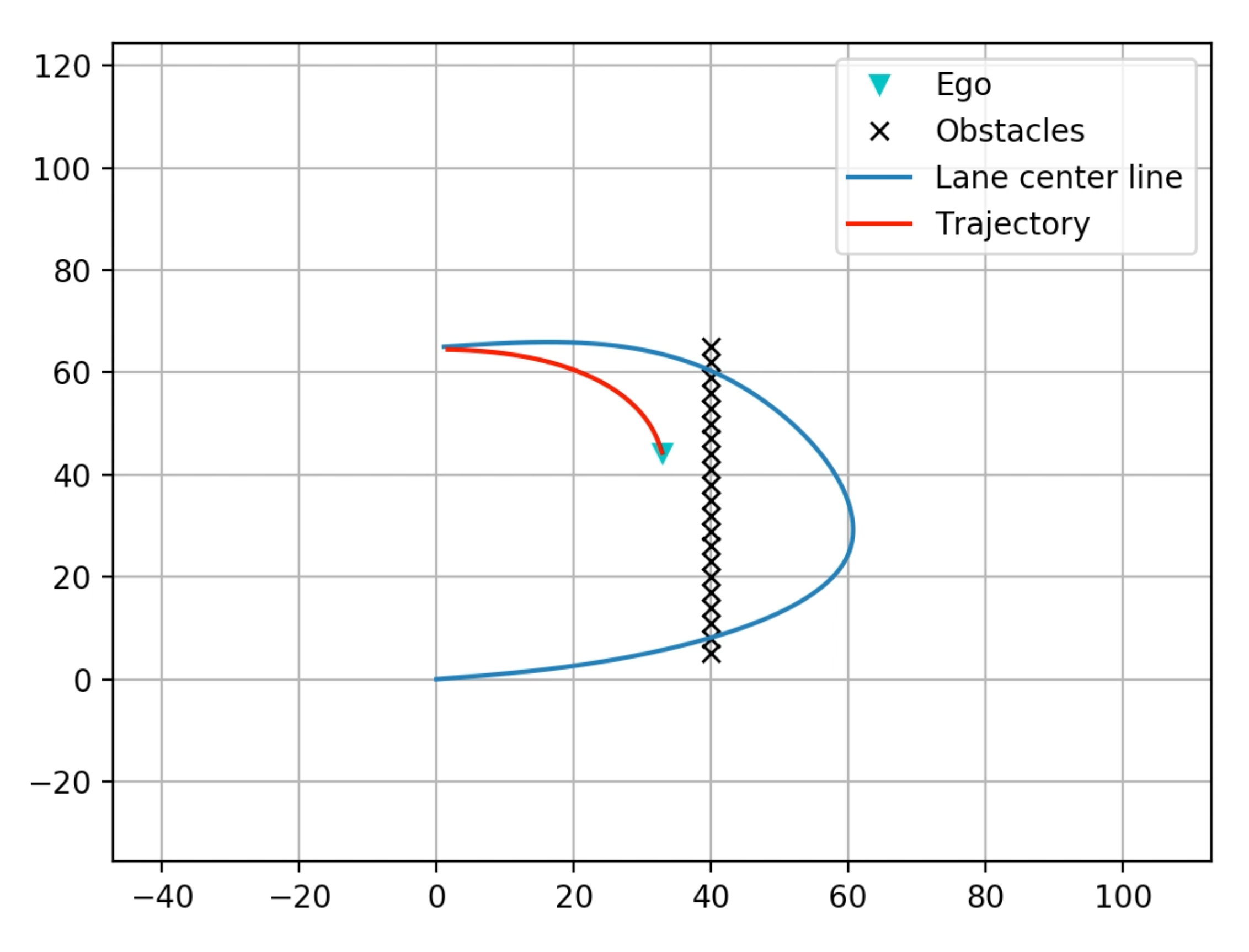} }
\subfigure[]{\includegraphics[width=4.1cm]{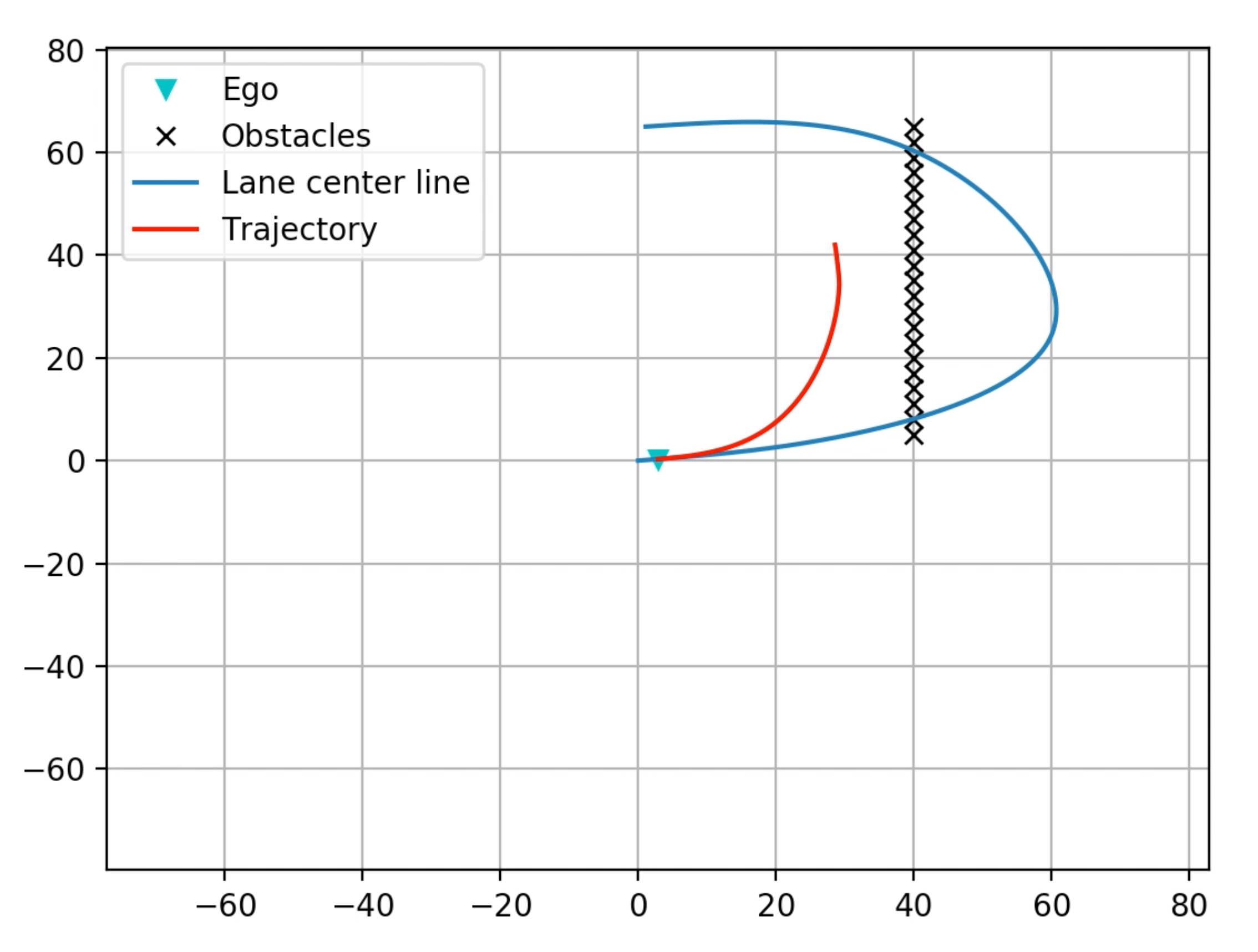} }
\subfigure[]{\includegraphics[width=4.1cm]{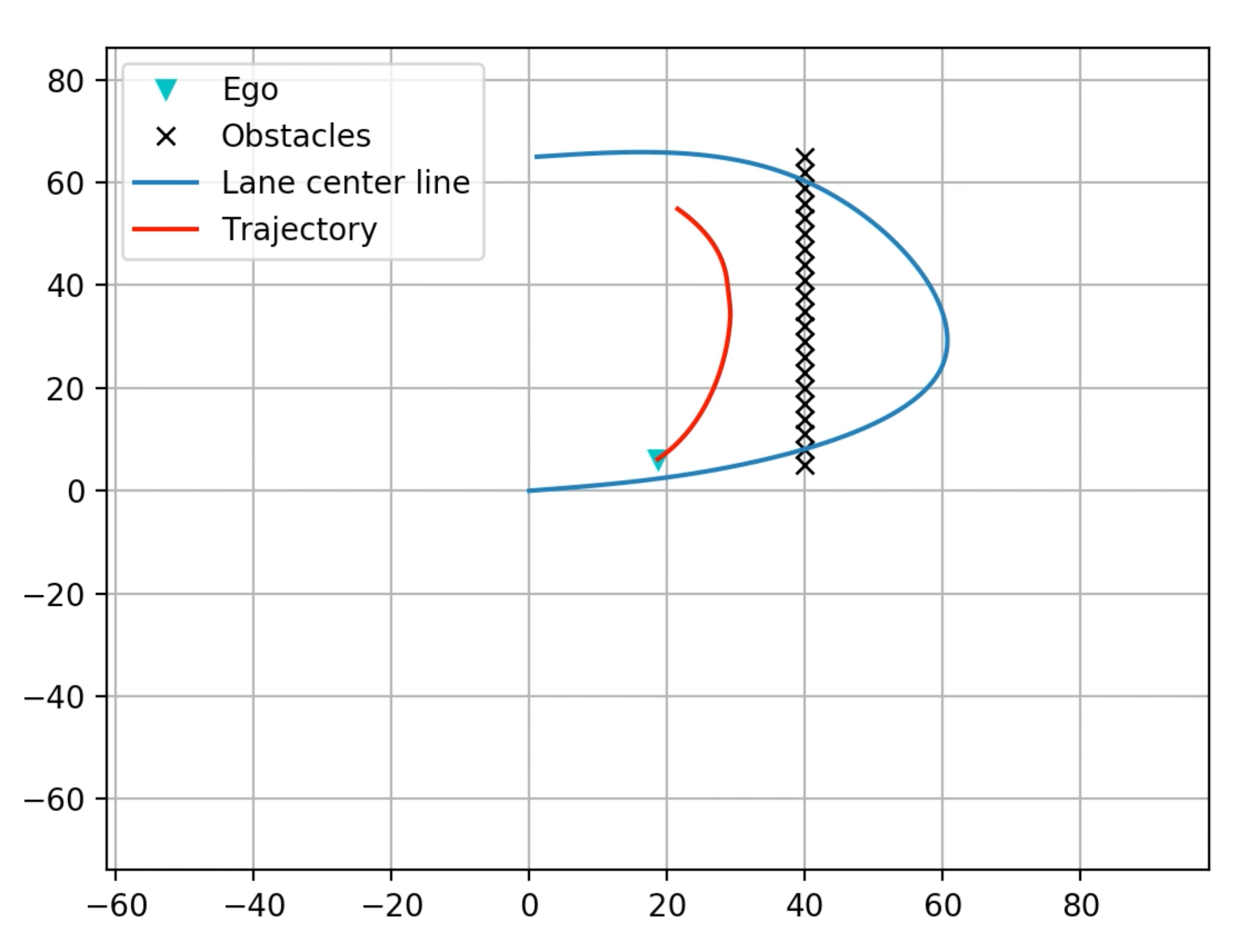} }
\subfigure[]{\includegraphics[width=4.1cm]{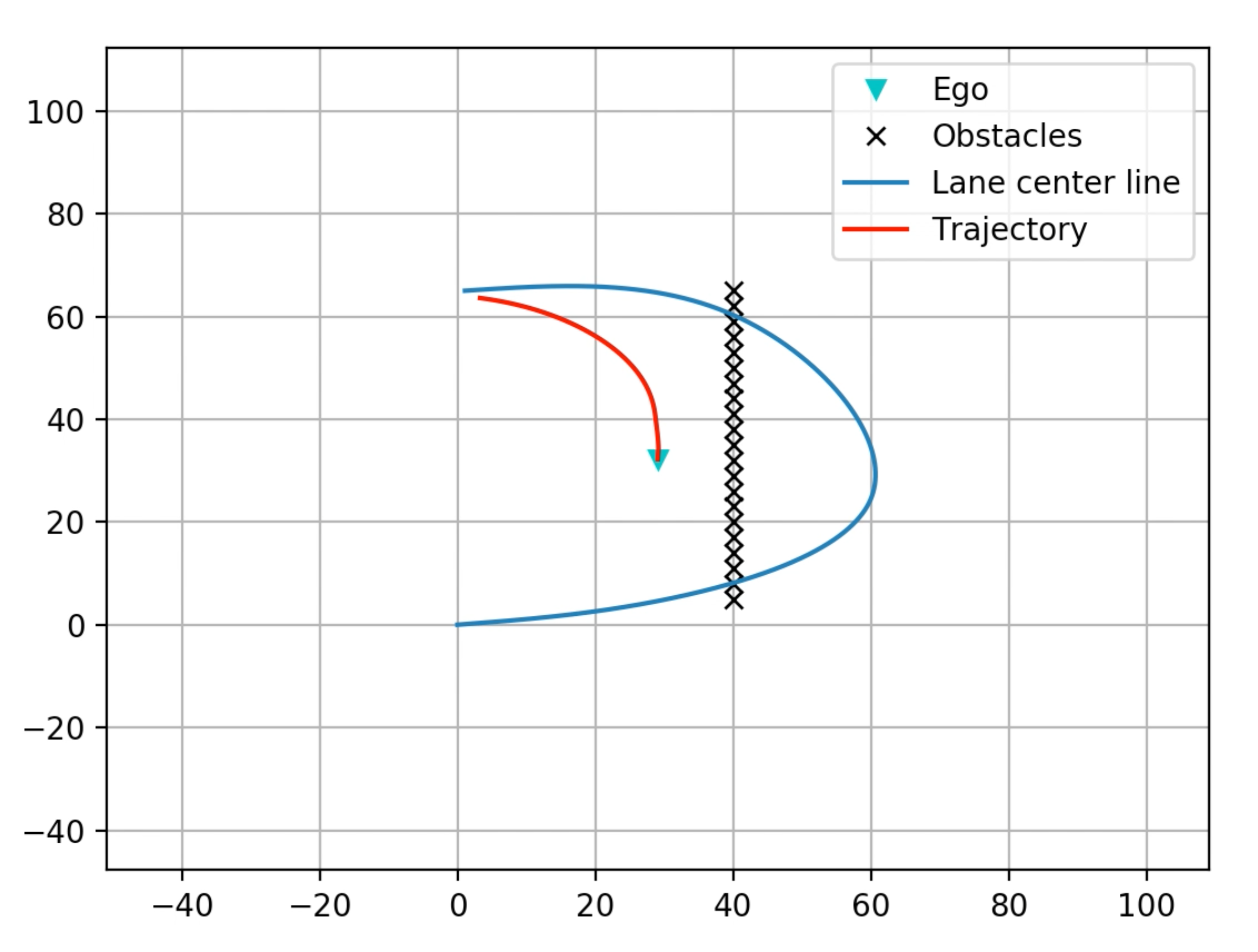} }
\subfigure[]{\includegraphics[width=4.1cm]{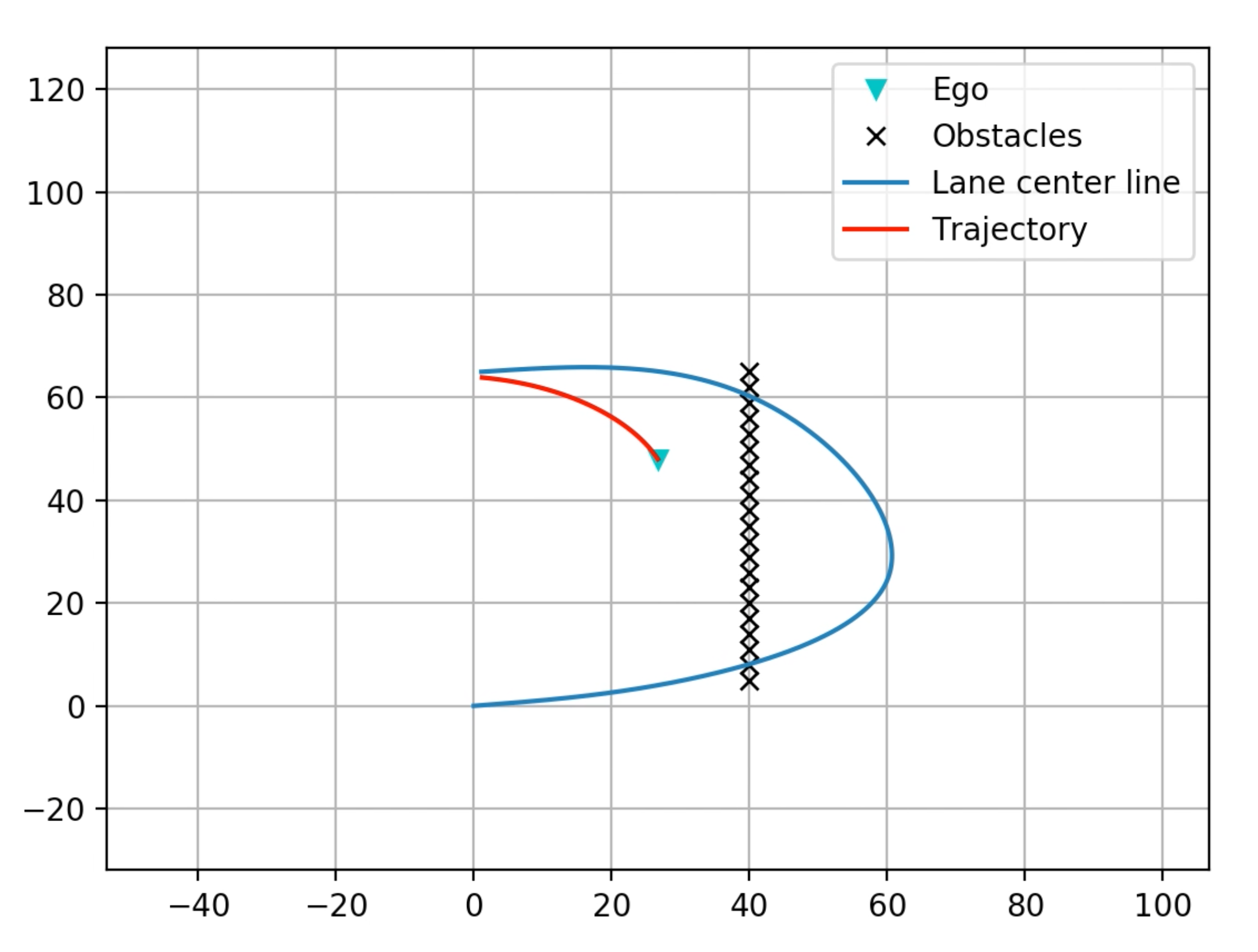} }
\caption{The sub-figures shown in (a)-(d) illustrate a result of trajectory planning by Algorithm \ref{alg4} . The sub-figures shown in (e)-(h) present a result of  trajectory planning by  Algorithm \ref{alg5}.} \label{fig4}
\end{figure*} 

In this section, we create an artificial environment for trajectory planning.  The generation algorithms \ref{alg4} and \ref{alg5} are used and compared in trajectory planning. High resolution video clips of these experimental results have been uploaded as supplement materials.


\subsection{Experimental settings}

In our planning environment, an U shaped curved road is assumed and the lane-center curve is provided.  Additionally, a static obstacle is placed near the corner of the road to mimic parking cars or road blockers. A moving agent is required to move from one end point of the road center to the other end point. The moving agent should follow the lane-center curve smoothly and avoid the obstacle.  

The detailed trajectory planning consists of the following steps:
\begin{itemize}
\item Implement candidate trajectory generation method multiple times to create many candidate trajectories. ({\bf Algorithm \ref{alg4}} or {\bf Algorithm \ref{alg5}})
\item Use the constrained planning optimizer to generate the final planed trajectory. 
\end{itemize}

\subsection{Experimental results and discussions}

We recorded the whole simulation process of the planning and movement of the moving agent in the form of videos. Screenshots of several representative time stamps from the videos are from collected and shown in Fig.  \ref{fig4}.  The blue curve denotes the lane center line and also the reference curve. The moving agent is represented as a triangle and its planned trajectory is shown as the red line. The obstacle is denoted as black cross marks. 

Figs. \ref{fig4}(a)-(e) are the screenshots from the experimental results based on Algorithm 4 for trajectory generation. 
Fig.   \ref{fig4}(a) is the initial state, where the planned trajectory works well. However, in Fig. \ref{fig4}(b),  the planned trajectory is heading the opposite with the reference curve. The reason why such a situation occurs is that at this moment there exists multiple points on reference curve (lane center line) which have the same nearest distance to the last point of trajectory. Since {\bf Algorithm \ref{alg4}} randomly samples the points from the reference curve, it may choose a wrong projection point and thus follow the opposite direction to the reference curve.  Figs. \ref{fig4}(c) and (d) are the consequences of this issue when the moving agent continues to move forward. Apparently, the generated  trajectories is unnatural and unacceptable for a car to execute.  Actually, the issue occurs frequently and may cause serious accidents when an autonomous vehicle passing a U-shaped curved road.

For comparisons, the results of trajectory planning based on {\bf Algorithm\ref{alg5}} have been shown in Fig. \ref{fig4}(f)-(g).  Fig. \ref{fig4}(f) is the initial state of the moving agent and the planned trajectory is more reasonable than the one shown in Fig. \ref{fig4}(a). Then, instead of randomly picking a number,  {\bf Algorithm \ref{alg5}} decides whether the requirement of product of$\kappa$ and lateral distance is smaller than 1 is successfully satisfied. If not, it uses {\bf Algorithm \ref{alg1}} to calculate the Frenet coordinate. Subsequently, {\bf Algorithm \ref{alg5}} manages to address the problem caused by {\bf Algorithm \ref{alg4}} by choosing the biggest index among those nearest points. As a result, the Fig. \ref{fig4}(f) shows a more reasonable planned trajectory. Fig. \ref{fig4}(g) and (h) are the following steps of the moving agent which successfully passes the obstacle and follows the U-shaped curved road.

\section{CONCLUSION}
Classic trajectory planning algorithms with Frenet frame make a common implicit assumption that any segment on the reference curve is the image (math) of some part of the planned trajectory, which means the agent should  follow the reference curve continuously. However, this assumption is not always true in practice and it may cause some serious issues.

In this paper, we study and analyze these issues and discussed their possible consequences in details. Then we propose an efficient candidate trajectory generation algorithm where a new projection operator from the Cartesian frame to Frenet frame is used. The new trajectory generation method is able to generate as many as possible candidate trajectories that follows the reference curve at the same direction for subsequent planning and optimization. As far as we know, it is the first time that a trajectory generation method is proposed with the constrain of $\kappa d < 1$. The proposed  trajectory planning algorithm enjoys the same computational time complexity as the classic algorithms and has been successfully applied in our real practice.

\bibliographystyle{IEEEtran}
\bibliography{reference}

\end{document}